\pdfoutput=1

\documentclass[11pt]{article}
\usepackage{authblk}

\usepackage[]{acl}

\usepackage{times}
\usepackage{latexsym}

\usepackage[T1]{fontenc}

\usepackage[utf8]{inputenc}

\usepackage{microtype}
\usepackage{graphicx}
\usepackage{booktabs}

\usepackage{todonotes}

\usepackage{microtype}

\title{What do Language Models know about word senses? \\Zero-Shot WSD with Language Models and Domain Inventories}

\author[ ]{Oscar Sainz}
\author[ ]{Oier Lopez de Lacalle}
\author[ ]{Eneko Agirre}
\author[ ]{German Rigau}

\affil[ ]{HiTZ Basque Center for Language Technologies - Ixa NLP Group}
\affil[ ]{University of the Basque Country UPV/EHU}
\affil[ ]{\{oscar.sainz, oier.lopezdelacalle, e.agirre, german.rigau\}@ehu.eus}

\begin{document}
\maketitle
\begin{abstract}
Language Models are the core for almost any Natural Language Processing system nowadays. One of their particularities is their contextualized representations, a game changer feature when a disambiguation between word senses is necessary. In this paper we aim to explore to what extent language models are capable of discerning among senses at inference time. We performed this analysis by prompting commonly used Languages Models such as BERT or RoBERTa to perform the task of Word Sense Disambiguation (WSD). 
We leverage the relation between word senses and domains, and cast WSD as a textual entailment problem, where the different hypothesis refer to the domains of the word senses. Our results show that this approach is indeed effective, close to supervised systems.

\end{abstract}

\section{Introduction}

It is undeniable that Language Models (LM) have drastically changed the Natural Language Processing (NLP) field~\cite{min2021recent}. More recently, those LM have also shown to be capable of performing NLP tasks with just few examples given in the context~\cite{gpt-3}, using the so called \textit{prompting}. One of their particularities, and the key difference with previous approaches, is their contextualized token representation. Allowing the model to adopt different representations for words (tokens) depending on the context has supposed a huge advantage when sense disambiguation is required for a given inference. But, \textbf{to what extent do LM actually know about word senses?} In this work, we tried to answer that question by evaluating LMs directly on the Word Sense Disambiguation (WSD) task via prompting.

Word Sense Disambiguation is the task of identifying the correct sense of a word in a given context. Current state-of-the-art on WSD involves fine-tuning a LM on SemCor~\cite{miller-etal-1994-using} to predict the correct among all possible sense glosses of the word in the given context. Other methods leverage the contextual representations of LM to perform WSD with a simple K-NN algorithm on the embedding space. Lately, the use of domain inventories was proposed to alleviate the high granularity of knowledge-bases~\cite{csi}. Recent studies that worked on zero-shot WSD refer to the task of predicting the senses of new lemmas not seeing during training as zero-shot~\cite{csi} WSD, however we aim for a completely zero-shot evaluation, where no annotated data is available for any lemma.

\begin{figure}
    \centering
    \resizebox{0.95\linewidth}{!}{
        \includegraphics{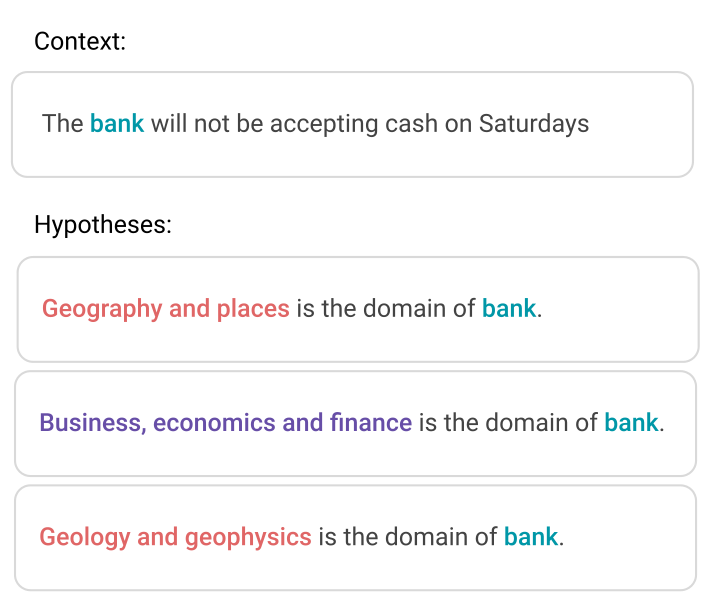}
    }
    \caption{An example of the Word Sense Disambiguation task converted to Textual Entailment, where the hypothesis refer to the possible domains of word senses. To solve the task a model would be asked to select the most probable hypothesis based on the context.}
    \label{fig:example}
\end{figure}

\begin{figure*}
    \centering
    \resizebox{\linewidth}{!}{
    \includegraphics{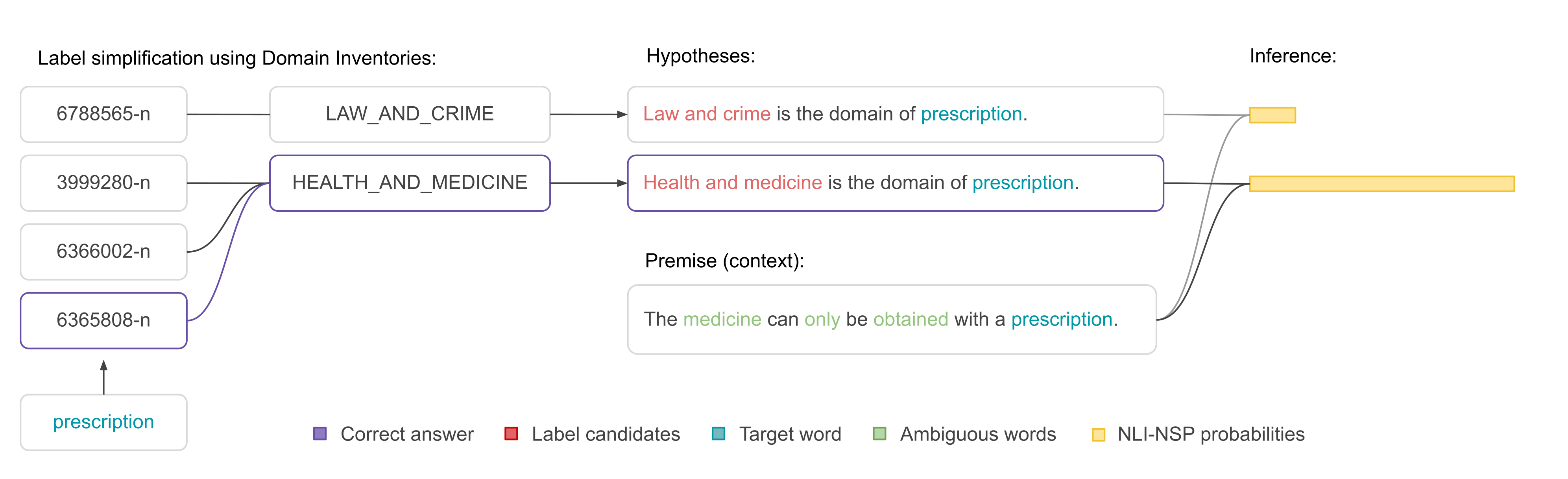}
    }
    \caption{Graphical description of the zero-shot WSD approach using Domain Inventories.}
    \label{fig:inference}
\end{figure*}

Despite the knowledge already encoded in the LM, training data is used in one way or another to introduce knowledge about the task. To avoid drawing noisy conclusions, we evaluated the LM as they are, without further fine-tuning on or using any kind of WSD training data. To that end, we prompted LMs like BERT~\cite{devlin-etal-2019-bert} and RoBERTa~\cite{liu2019roberta} to perform a task that requires WSD knowledge to be successfully solved. 

Figure~\ref{fig:example} shows an example of how a model can be prompted to solve WSD using Textual Entailment as a proxy. On this example we consider that the word bank has senses from three different domains: \textit{Geography and places}, \textit{Business, economics and finance} and \textit{Geology and geophysics}. The three possible domains are converted to hypothesis using predefined prompts. Finally, a supervised Textual Entailment model is used to perform the inference. More details on of the approach are discussed in Section~\ref{sec:prompting}.

In this work we first evaluated commonly used LMs as a zero-shot domain labelers with 3 different domain inventories. Then, following~\cite{csi} we addressed the WSD using domain inventories and evaluated the LMs on them. We showed that LMs have some notion of senses as they perform zero-shot WSD significantly better than a random baseline and sometimes close to the supervised state-of-the-art. We also provided different analysis comparing different prompts and performing an error analysis over the two evaluated tasks.

\section{Prompting Language Models} \label{sec:prompting}

Since the past few years, prompting has become the \textit{de facto} approach to probe language models ~\cite{li-etal-2022-probing-via}. ~\citet{min2021recent} defined prompting as the practice of adding natural language text, often short phrases, to the input or output to encourage pre-trained models to perform specific tasks. However, due to its wide definition, several different ways of prompting exists, such as \textit{instruction based}, \textit{template-based} or \textit{proxy-task based}. For more information about prompting we encourage the reader to read the \citet{10.1145/3560815} survey.

In this work we focused on the \textit{proxy-task based} approach, more precisely, we made use of the Next Sentence Prediction (NSP) and Textual Entailment (TE) tasks as a proxy. The TE is also known as Natural Language Inference (NLI), we will use both terms interchangeably.  The choice of this approach was made based on previous works on zero-shot domain labelling ~\cite{sainz-rigau-2021-ask2transformers}.

Both, NSP and TE are sentence-pair classification tasks: the first attempts to predict whether a sentence is followed by another and the second aims to predict if an entailment relation exists between both sentences (premise and hypothesis).
Figure~\ref{fig:inference} shows an example of how to perform WSD using NSP or TE models. The process can be briefly summarized as follows: (1) for each possible sense $s$ of the target word $w$ we obtain their corresponding domain $d$ using a domain inventory $D$ (domain inventories are discussed in more detail in Section~\ref{sec:inventories}). (2) predefined prompts are used to generate verbalizations that will serve as possible continuations (on NSP) or hypothesis (on TE) $h$. (3) a pretrained NSP or TE model is used to obtain a probability for each sentence/hypothesis and therefore, to each domain. Formally, for a TE model we defined the probability of word $w$ being from domain $d_i \in D^w$ in context $c$ as follows:

\begin{equation}
    P(d_i|c, w) = P(\textnormal{entailment} | c, h_{wi})
\end{equation}
\noindent where $h_{wi}$ is the hypothesis generated using a predefined prompt, the domain label $d_i$ and the word $w$. Similarly, for a NSP model the probability is defined as follows:

\begin{equation}
    P(d_i|c, w) = P(\textnormal{is\_next} | c, h_{wi})
\end{equation}

\begin{table*}
    \centering
    \resizebox{\textwidth}{!}{
        \begin{tabular}{r|p{0.15\textwidth}p{0.2\textwidth}p{0.15\textwidth}|p{0.4\textwidth}}
            \toprule
            Sense & BabelDomains & CSI & WN Domains & Gloss \\
            \midrule
            00006484-n & Biology & Biology & biology & The basic structural and functional unit of all organisms; ...\\
            \midrule
            02991048-n & Chemistry and mineralogy & Craft, Engineering and Technology & electronics & A device that delivers an electric current as the result of a chemical reaction. \\
            \midrule
            02992529-n & Computing & Craft, Engineering and Technology & electricity \quad telephony & A hand-held mobile radiotelephone for use in an area divided into small sections, each with its own short-range transmitter/receiver \\
            \bottomrule
        \end{tabular}
    }
    \caption{Example of Domain inventories for 3 senses of the word \textit{cell}.}
    \label{tab:domains_example}
\end{table*}

\noindent Table ~\ref{tab:prompts} shows the prompts used for probing Language Models in Domain Labelling and Word Sense Disambiguation tasks.


\section{Domain Inventories} \label{sec:inventories}

A domain inventory is a set of domain labels such as \textit{Health and Medicine}, \textit{Culture} or \textit{Business and economics} that aims to cover the wider spectrum of domains as possible with a specific granularity level. Actually, these domain inventories are used to label synsets from knowledge-bases like WordNet~\cite{wordnet} and BabelNet~\cite{babelnet}. Examples of WordNet synset annotations from different domain inventories are shown in the Table~\ref{tab:domains_example}. Recent studies ~\cite{csi} suggest to use domain inventories to address the high granularity problem that affects WSD tasks. In this section we describe the three domain inventories on which we evaluated the Language Models.

\paragraph{BabelDomains}~\cite{camacho-collados-navigli-2017-babeldomains} is a unified resource that includes domain information for Wikipedia, WordNet and BabelNet. It inherits the domains from Wikipedia domains of knowledge, a total of 34 coarse labels. Although it is semi-automatically annotated, two gold standard datasets (for WordNet and Wikipedia) are provided for evaluation.

\paragraph{Coarse Sense Inventory (CSI)}~\cite{csi} was created to reduce the level of granularity of WordNet synsets while maintaining their expressiveness. It contains a total of 45 labels shared across the lexicon. Compared to previous alternatives, CSI provided a higher agreement among annotators. Also it was already proven to be useful for the WSD task.

\paragraph{WordNet Domains}\cite{bentivogli-etal-2004-revising} is a fine-grained domain inventory containing about 160 labels. It is organised in a hierarchical way, from global concepts such as \textit{pure\_science} to specific concepts as \textit{oceanography}. This inventory provides a domain label to each synset in WordNet. Due to the hierarchical nature and fine granularity, in our experiments we kept only the domain labels until the third level, mapping all the labels below to the closest available domain. We end up with 60 domain labels.

\section{Experimental Setup}

\begin{table*}
    \centering
    \resizebox{0.9\textwidth}{!}{
        \begin{tabular}{r|l}
            \toprule
            Task & Prompt \\
            \midrule
            Domain Labelling & \{gloss\} | The domain of the sentence is about \{label\}. \\
            Word Sense Disambiguation & \{context\} | The domain of the sentence is about \{label\}. \\
             &  \{context\} | \{label\} is the domain of \{word\}. \\
            \bottomrule
        \end{tabular}
    }
    \caption{Prompts used for probing Language Models.}
    \label{tab:prompts}
\end{table*}

In this section we describe the models we evaluated, and the Domain Labelling and Word Sense Disambiguation tasks we used for evaluation.

\paragraph{Models.} For the experiments we decided to evaluate two very commonly used models: BERT and RoBERTa. We followed previous works on zero-shot domain labelling~\cite{sainz-rigau-2021-ask2transformers} for approach and model selection. As explained in Section~\ref{sec:prompting} we required that the models were already fine-tuned to perform sentence pair classifications. In the case of the BERT models, we used the LM itself with the NSP head that was trained during pre-training, in the tables it is shown as NSP. For the case of RoBERTa, as it has not been pre-trained for any sentence classification task, we evaluated two checkpoints that were also fine-tuned with TE data: NLI and NLI*. The main difference between both checkpoints is the variety of data on which the models were trained. We evaluated the \textit{large} variant of those models. The NLI variation was trained just on MultiNLI~\cite{williams-etal-2018-broad} dataset and NLI* variations was also trained on SNLI~\cite{bowman-etal-2015-large}, Fever-NLI~\cite{thorne-etal-2018-fever} and Adversarial-NLI~\cite{nie-etal-2020-adversarial}. Both models are publicly available at HuggingFace Model Hub~\cite{wolf-etal-2020-transformers}.

\paragraph{Domain Labelling task} is the task of classifying some text $t$ into a set of domain labels $D$. In our case, the text to classify are WordNet synset glosses and the domain labels are the ones defined by the domain inventories. The task was evaluated on a small manually annotated dataset released by \citet{camacho-collados-navigli-2017-babeldomains}. The dataset consist of domain annotations for 1540 WordNet synsets using BabelDomains inventory. For those 1540 synsets we also collected the domain information from CSI and WordNet Domains. The 3 checkpoints described above were evaluated with each domain inventory. To evaluate the models on domain labelling data we used the prompts described in Table~\ref{tab:prompts} to convert domain labelling examples into NLI or NSP examples. The prompt is used to generated as many hypotheses as labels are in the inventory, by replacing the \textit{gloss} placeholder with the synset's gloss and the \textit{label} placeholder with the corresponding label each time. 

\begin{figure}[!ht]
    \centering
    \begin{quote}
        	Cell: \textbf{(biology)} the basic structural and functional unit of all organisms; ...
    \end{quote}
    \caption{An example of WordNet gloss. The hint in the gloss is highlighted.}
    \label{fig:gloss_example}
\end{figure}

WordNet glosses sometimes contains domain information inside them. For example, in the gloss shown in Figure~\ref{fig:gloss_example} the domain information is highlighted in bold. We will call them domain \textit{hints}. As we are using those glosses as inputs to predict the domain of the synsets, the hints give a huge advantage to the models. Therefore, for the evaluation we considered two alternatives: with and without hints.

\paragraph{WSD task} is the task of identifying the correct sense $s$ a word $w$ withing a context $c$ among all its possible senses $s \in S^w$. In this case, and following recent works we reframed the task from predicting senses to more coarse set of labels (domains)~\cite{csi}. Therefore, the task aims to classify the domain of the correct sense $d_s$ among the domains of the possible senses $D^w$. As senses in WordNet are very fine-grained, several senses of the same domain may coexist, after replacing them with their domain the set of possible labels might be reduced, therefore $|D^w| \leq |S^w|$. An example of two senses from the same domain is shown in Table~\ref{fig:gloss_example}. The task was evaluated on the standard commonly known SemEval~\cite{pradhan-etal-2007-semeval, navigli-etal-2013-semeval, moro-navigli-2015-semeval} and Senseval~\cite{edmonds-cotton-2001-senseval, snyder-palmer-2004-english} datasets. For each model, we also compared two different prompts shown in Table~\ref{tab:prompts}: the first is the same as the one used for Domain Labelling and is used to predict the domain of the whole context; the second instead adds a reference to the target word, and is intended to focus the model to predict the domain of the given word withing the context. Finally, we report a random guessing baseline and a supervised upper-bound from ~\citet{csi}.

\section{Results}

\begin{figure*}
    \centering
    \resizebox{0.7\textwidth}{!}{
        \includegraphics{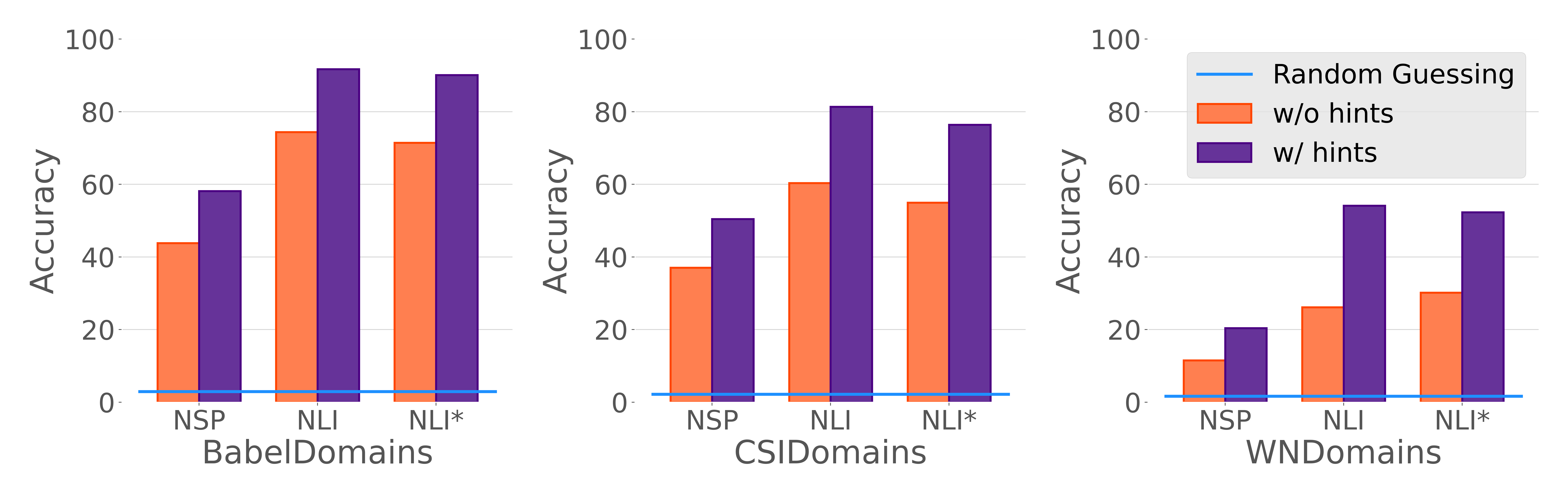}
    }
    \caption{Results on Domain Labelling task for three different domain inventories.}
    \label{fig:topic_results}
\end{figure*}

In this section we discuss the results obtained on each experiment. First we discuss the results obtained on the Domain Labelling task. Then, we show the results from Word Sense Disambiguation. And finally we analyze the correlation between both tasks as they share the label space.

\paragraph{Are Language Models able to discriminate domains in sense glosses?} Figure~\ref{fig:topic_results} shows the results obtained for the Domain Labelling task. As a general overview, the three models obtain decent results considering no data for training was provided. Comparing NLI models vs the NSP model, we can conclude that NLI based models perform better in all cases, in concordance with previous works~\cite{wang2021entailment}. However, additional TE data (NLI vs NLI*) does not seem to be very useful for the task. Finally, the results shows that the domain hints in the gloss affects significantly to the performance, specially in WordNet Domains, where the labels are very fine-grained.

\begin{figure*}[!ht]
    \centering
    \resizebox{1.\textwidth}{!}{
        \includegraphics{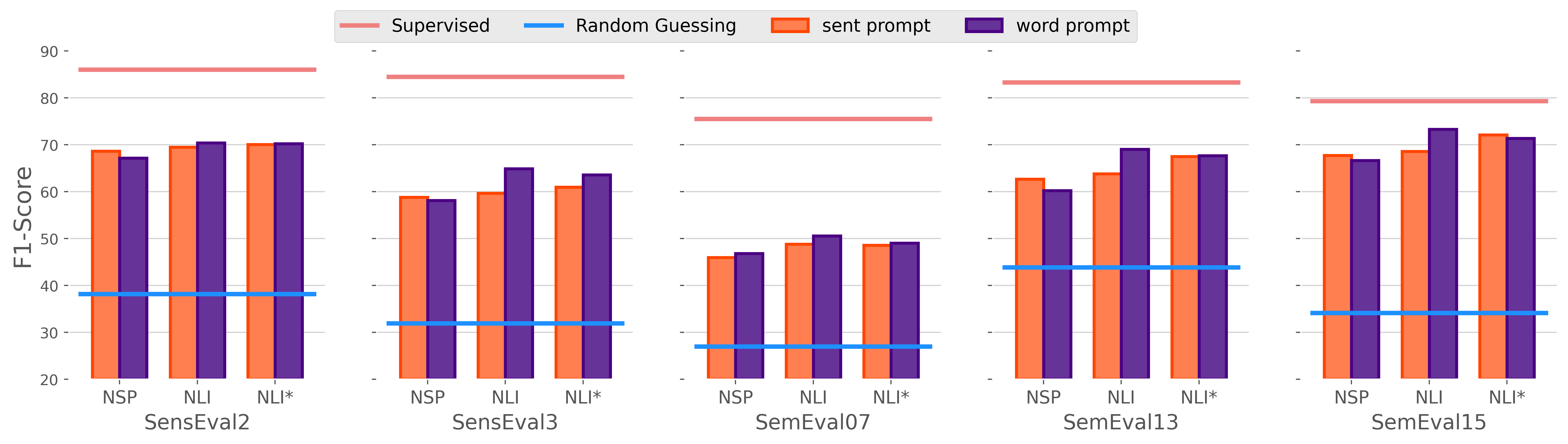}
    }
    \caption{Word Sense Disambiguation results for the three systems in the 5 evaluation datasets. The red line indicates the state-of-the-art supervised scores and the blue line the scores obtained by random guessing.}
    \label{fig:wsd_results}
\end{figure*}

\paragraph{Do Language Models know about Word Senses?} Figure~\ref{fig:wsd_results} shows the results for each of the WSD datasets along with random and supervised baselines. In general, the results suggest that \textbf{in fact the Language Models know about senses}. While still far from a supervised upper-bound, the three models have shown significantly better performance than a random classifier. Moreover, for the SemEval-15 task the models achieve a performance close to the upper-bound. Comparing the NSP model against the NLI models, the same pattern as in the Domain Labelling task occur, the NLI models are better in all scenarios. If we compare both TE models, both perform similarly when the \textit{sentence prompt} is used, for the \textit{word prompt} instead the NLI model shows slightly better results. Overall, the best combination is NLI model with the \textit{word prompt}.

\begin{table}
    \centering
    \resizebox{.95\linewidth}{!}{
    \begin{tabular}{r|llll|l}
        \toprule
        Model & Noun & Adj & Verb & Adv & All \\
        \midrule
        Random & 40.7 & 48.4 & 23.7 & 59.1 & 38.8 \\
        \midrule
        \multicolumn{6}{c}{\textit{Sentence prompt}} \\
        \midrule
        NSP & 60.3 & 84.9 & 50.4 & 86.6 & 62.6 \\
        NLI & 64.3 & 86.2 & 54.8 & 86.4 & 66.1 \\
        NLI* & 65.0 & 85.9 & 55.0 & 85.3 & 66.4 \\
        \midrule
        \multicolumn{6}{c}{\textit{Word prompt}} \\
        \midrule
        NSP & 59.4 & 84.8 & 50.2 & 86.4 & 61.9 \\
        NLI & \textbf{66.2} & \textbf{86.8} & \textbf{57.0} & \textbf{87.3} & \textbf{67.8} \\
        NLI* & 65.3 & 85.5 & 55.7 & 85.5 & 66.8 \\
        \bottomrule
    \end{tabular}
    }
    \caption{F1-Scores per word category}
    \label{tab:word_category}
\end{table}

\paragraph{Do Language Models perform differently depending on the word category?} To answer this question we report the results grouped by the word category in the Table~\ref{tab:word_category}. The table reports the same results as Figure~\ref{fig:wsd_results} except for the supervised upper-bound which has not been reported by \citet{csi} under this setting. We also report the \textit{micro-averaged} F1-Score for all categories, allowing us to clearly compare all the systems. Considering the results, the NLI model with the \textit{word prompt} is again the best performing system across all word categories. Comparing the NLI\textsubscript{word} model against the random baseline we can observe a high correlation in the scores, which suggest that the errors on each category depend more on the task difficulty rather than specific language model issues.

\begin{figure}
    \centering
    \resizebox{1.05\linewidth}{!}{
        \includegraphics{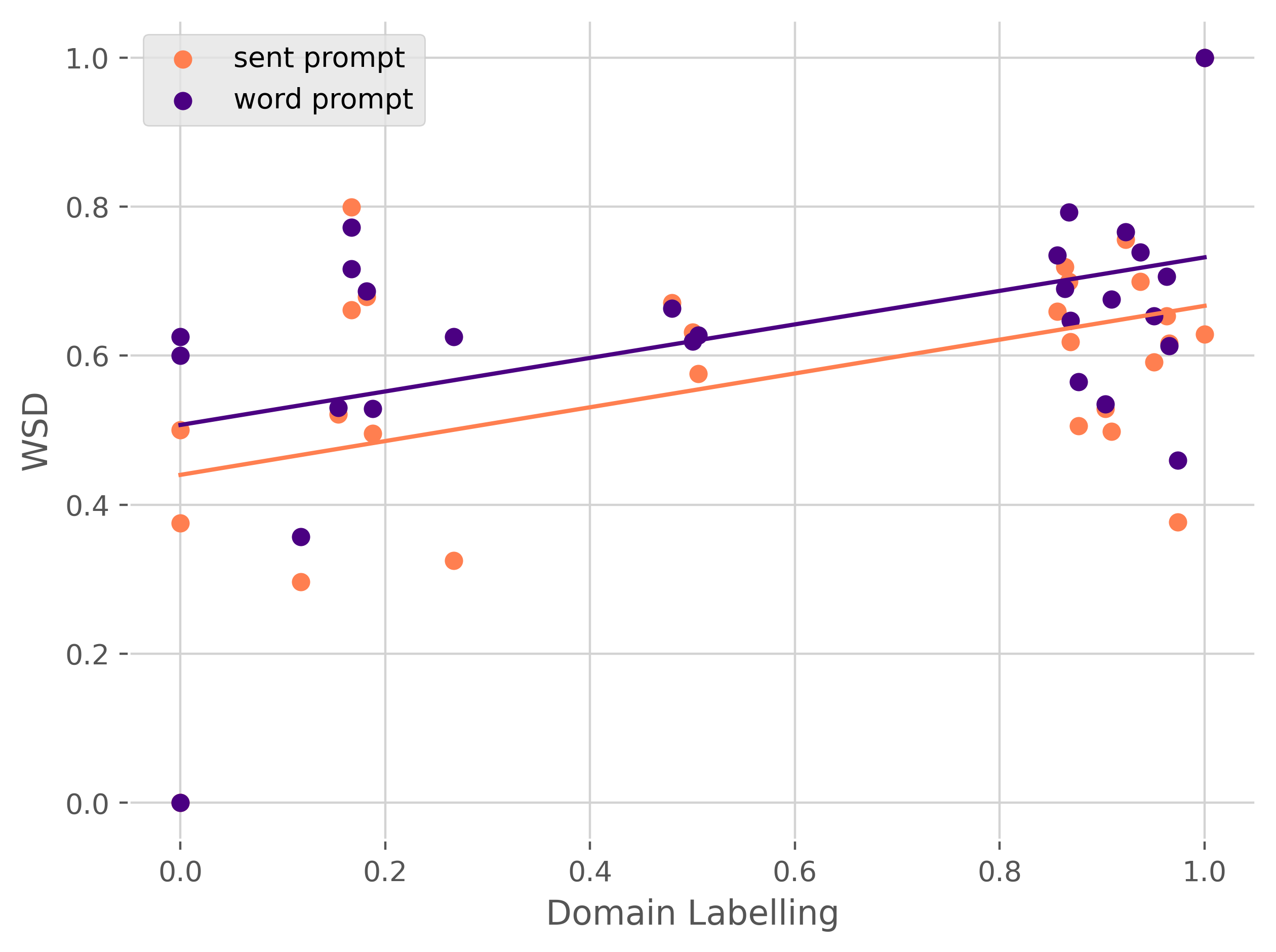}
    }
    \caption{F1 correlation between Domain Labelling and WSD tasks.}
    \label{fig:error-correlation}
\end{figure}

\begin{table}
    \centering
    \resizebox{1.0\linewidth}{!}{
    \begin{tabular}{r|lll}
        \toprule
         & Dom Lab. & WSD\textsubscript{sent} & WSD\textsubscript{word} \\
        \midrule
        Dom Lab. & 1.00 & 0.32 & 0.41 \\
        WSD\textsubscript{sent} & 0.32 & 1.00 & 0.81 \\
        WSD\textsubscript{word} & 0.41 & 0.81 & 1.00\\
        \bottomrule
    \end{tabular}
    }
    \caption{Spearman's correlation of F1-Scores between tasks using shared labels. The scores correspond to the NLI model.}
    \label{tab:taskwise_correlation}
\end{table}

\paragraph{To what extent does the performance on Domain Labelling affects WSD?} As we are framing WSD as a Domain Labelling problem, it is intuitive to think that the performance on Domain Labelling can affect the performance on WSD. The evaluation we carried out in both tasks have a common label space, and therefore, we can compute the correlation between label scores. For each label, we compared the F1-score obtained on Domain Labelling and WSD tasks. Figure~\ref{fig:error-correlation} shows the per-domain F1 scores on Domain Labelling and WSD tasks, each point represents the F1 obtained on a specific label. In the figure, we included the F1 for both \textit{sentence prompt} and \textit{word prompt} systems. The results shows \textbf{very little correlation} between both tasks. The Table~\ref{tab:taskwise_correlation} shows the Spearnman's correlation for each task pair. The results again shows that both tasks are poorly correlated, even when we use the same prompt. However, this comparison might not be completely fair, there are 2 main reasons that could affect the results: the Domain Labelling glosses have a particular structure and different from WSD contexts, also, on WSD the system needs to predict the correct among \textbf{possible} labels rather than all the label space as in Domain Labelling. We should take into consideration those differences at the time of interpreting the results.

\section{Related Work}


\paragraph{Word Sense Disambiguation} Approaches to WSD range from supervised that only use annotated data~\cite{agirre-etal-2014-random, hadiwinoto-etal-2019-improved, bevilacqua-navigli-2019-quasi} to knowledge-based~\cite{moro-etal-2014-entity, agirre-etal-2014-random, scozzafava-etal-2020-personalized}, as well as approaches that combine supervised and knowledge-based approaches~\cite{kumar-etal-2019-zero, bevilacqua-navigli-2020-breaking, blevins-zettlemoyer-2020-moving, conia-navigli-2021-framing, barba-etal-2021-esc}.

Knowledge-based approaches employ graph algorithms on a semantic network~\cite{moro-etal-2014-entity, agirre-etal-2014-random, scozzafava-etal-2020-personalized}, in which senses are connected through semantic relations and are described with definitions and usage examples. Unfortunately, their independence from annotated data comes at the expense of performing worse than supervised models~\cite{pilehvar-navigli-2014-large}.

Supervised approaches frame the task as a classification problem and use available annotated data to learn mapping the words in context to senses. Before supervised neural models emerged as state of the art in NLP, the task of supervised WSD was performed based on a variety of lexico-syntantic and semantic feature representations that are fed to a supervised machine learning classifier~\cite{zhong-ng-2010-makes}. Instead, current state-of-the-art  supervised models rely on the use of pretrained Transformers as core architecture of the model.  Among these models we can find approaches that exclusively use annotated data to learn effective representations of the target word in context and feed it to some classification head~\cite{raganato-etal-2017-neural, hadiwinoto-etal-2019-improved,bevilacqua-navigli-2019-quasi, conia-navigli-2021-framing}. 

Some approaches have shown that an effective way to improve sense representation is to exploit the glosses provided by the sense inventories. Gloss representation are  then incorporated to the sense embedding~\cite{peters-etal-2018-deep}, in which  the most probable sense is retrieve according to the similarity with the given context. Multiple works have been shown effective in WSD such as LMSS~\cite{loureiro-jorge-2019-language}, SensEmBERT~\cite{Scarlini_Pasini_Navigli_2020}, ARES~\cite{scarlini-etal-2020-contexts}, SREF~\cite{wang-wang-2020-synset},  EWISE~\cite{kumar-etal-2019-zero} and  EWISER~\cite{bevilacqua-navigli-2020-breaking}, among many others.  Glosses have also been exploited in sequence-tagging approaches~\cite{huang-etal-2019-glossbert, yap-etal-2020-adapting}, where the task is framed as sequence classification problem~\cite{barba-etal-2021-esc}. In a similar manner,~\cite{bevilacqua-navigli-2020-breaking} propose a generative approach to cast WSD as sequence classification problem.  In adition to glosses, other approaches presented ways to make use of the knowledge encoded in KBs such as WordNet. For instance, \cite{loureiro-jorge-2019-language, wang-wang-2020-synset} propagate sense embeddings using WordNet as a graph.  Please refer to \cite{wsd-survey} to obtain further details of the recent trends in WSD.

\paragraph{Prompting Language Models} has changed the paradigm of how Language Models can be used to extract even more potential from them. Initially with very large LM like GPT-3~\cite{gpt-3} and later with smaller ones~\cite{gao-etal-2021-making} prompts allowed the models to perform zero or few-shot classifications with simple natural language. This ability also allowed models to improve performance on data-scarce problems by large margin \cite{le-scao-rush-2021-many,min2021recent,10.1145/3560815}. These prompts can be discrete \cite{gao-etal-2021-making, schick-schutze-2021-exploiting, schick-schutze-2021-shot, schick-schutze-2021-just} close to natural language or continuous \cite{liu-etal-2022-p} close to other efficient deep learning methods like Adapters~\cite{pfeiffer-etal-2020-adapterhub}. Closer to our work, Textual Entailment~\cite{10.1007/11736790_9} has been used as a source of external supervision to solve several text classification tasks \cite{yin-etal-2019-benchmarking, yin-etal-2020-universal, wang2021entailment, sainz-rigau-2021-ask2transformers, mccann2018natural, white-etal-2017-inference}, Named Entity Recognition~\cite{li-etal-2022-ultra, poliak-etal-2018-collecting, yang-etal-2022-see}, Relation Extraction~\cite{levy-etal-2017-zero, sainz-etal-2021-label}, Event Extraction~\cite{lyu-etal-2021-zero}, Event Argument Extraction~\cite{sainz-etal-2022-textual, sainz-etal-2022-zs4ie}, Intent Classification~\cite{xia-etal-2021-incremental}, Aspect-based Sentiment Analysis~\cite{zeroshotabs} and many more.


\paragraph{Domain Inventories.} 

Domain information was added to Princeton WordNet~\cite{wordnet} since version 3.0. In total 440 topics were represented as a synsets in the graph. The topic label assignment was achieved through pointers from source synsets to target synsets. Being the most frequent topic is \textsc{law, jurisprudence}. However, the manual assignment of topic labels to synsets in WordNet is very costly. As a consequence, semi-automatic methods were developed. For instance, WordNet Domains~\cite{bentivogli-etal-2004-revising} is a semi-automatically annotated domain inventory that labels WordNet synsets with 165 hierarchically organised domains. The use of domain inventories such as WordNet Domains, allowed to reduce polysemy degree of WordNet synsets by grouping those that belong to the same domain \cite{magnini2002role}. However, far from being perfect, many synsets were labelled as \textsc{factotum}, meaning that the synset cannot be labelled with a particular domain. Several works were proposed to improve WordNet Domains, such as eXtended WordNet Domains~\cite{gonzalez-agirre-etal-2012-proposal, gonzalez2012graph}, that applied graph-based methods to propagate the labels through the WordNet structure.

Domain information is not only available in WordNet, for example IATE\footnote{\url{http://iate.europa.eu/}} is a European Union inter-institutional terminology database. The domain labels of IATE are based on the Eurovoc thesaurus\footnote{\url{https://op.europa.eu/en/web/eu-vocabularies/th-dataset/-/resource/dataset/eurovoc}} and were introduced manually. More recently, several new domain inventories appeared, such as BabelDomains~\cite{camacho-collados-navigli-2017-babeldomains} or Coarse Sense Inventory~\cite{csi}.

\section{Conclusions}

In this work we present an evaluation approach to test Language Models on the tasks of Domain Labelling and Word Sense Disambiguation without annotated data requirements. For the WSD task we followed \citet{csi} to reduce the granularity level. Our results showed that the Language Models we tested here \textbf{have some notion of word senses}. They easily outperformed the baseline, and sometimes almost reached to supervised systems performance. In addition, our further analysis shows that there is very low error propagation from Domain Labelling to WSD as their errors are poorly correlated. For the future, we plan to evaluate larger Language Models on the task to try to understand to what extent scaling these LMs affects to sense recognition.

\section*{Acknowledgments}
Oscar is funded by a PhD grant from the Basque Government (PRE\_2020\_1\_0246). This work is based upon work partially supported via the IARPA  BETTER Program contract No. 2019-19051600006 (ODNI, IARPA), DeepKnowledge (PID2021-127777OB-C21) project funded by MCIN/AEI/10.13039/501100011033 and by FEDER Una manera de hacer Europa, AWARE (TED2021-131617B-I00) project funded by MCIN/AEI/10.13039/501100011033 and by European Union NextGeneration EU/ PRTR, and by the Basque Government (IXA excellence research group IT1570-22). 

\bibliography{anthology,custom}
\bibliographystyle{acl_natbib}

\end{document}